\documentclass[english,10pt,twocolumn,twoside]{IEEEtran}
\usepackage[T1]{fontenc}
\usepackage[latin9]{inputenc}
\usepackage{geometry}
\usepackage{babel}
\usepackage{amsthm}
\usepackage{amsmath}
\usepackage{amssymb}
\usepackage{graphicx}
\usepackage[unicode=true,pdfusetitle,
 bookmarks=true,bookmarksnumbered=false,bookmarksopen=false,
 breaklinks=false,pdfborder={0 0 0},backref=false,colorlinks=false]
 {hyperref}

\makeatletter

\providecommand{\tabularnewline}{\\}

\theoremstyle{plain}
\newtheorem{thm}{\protect\theoremname}

\usepackage{algorithm}
\usepackage{algorithmic}
\usepackage{url}
\usepackage{graphicx}

\makeatother

\providecommand{\theoremname}{Theorem}

\begin{document}

\title{Learning Dictionaries with Bounded Self-Coherence}

\author{Christian D. Sigg, \emph{Member, IEEE,} and Tomas Dikk, and Joachim
M. Buhmann, \emph{Senior Member, IEEE\thanks{Copyright (c) 2012 IEEE. Personal use of this material is permitted. However, permission to use this material for any other purposes must be obtained from the IEEE by sending a request to pubs-permissions@ieee.org.}\thanks{Manuscript received June 03, 2012; revised July 23, 2012; accepted September 28, 2012. Date of publication October 10, 2012. The associate editor coordinating the review of this manuscript and approving it for publication was Prof. Eric Moreau.}\thanks{Christian Sigg is with the Swiss Federal Office of Meteorology and Climatology (MeteoSwiss), Zurich, Switzerland (e-mail: christian@sigg-iten.ch).}\thanks{Tomas Dikk was with the Department of Computer Science, ETH Zurich, Switzerland and is now self-employed. (e-mail: \mbox{tomasdikk@tomasdikk.com}).}\thanks{Joachim Buhmann is with the Department of Computer Science, ETH Zurich, Switzerland (e-mail: jbuhmann@inf.ethz.ch).}\thanks{Digital Object Identifier 10.1109/LSP.2012.2223757}}}
\maketitle
\begin{abstract}
Sparse coding in learned dictionaries has been established as a successful
approach for signal denoising, source separation and solving inverse
problems in general. A dictionary learning method adapts an initial
dictionary to a particular signal class by iteratively computing an
approximate factorization of a training data matrix into a dictionary
and a sparse coding matrix. The learned dictionary is characterized
by two properties: the coherence of the dictionary to observations
of the signal class, and the self-coherence of the dictionary atoms.
A high coherence to the signal class enables the sparse coding of
signal observations with a small approximation error, while a low
self-coherence of the atoms guarantees atom recovery and a more rapid
residual error decay rate for the sparse coding algorithm. The two
goals of high signal coherence and low self-coherence are typically
in conflict, therefore one seeks a trade-off between them, depending
on the application. We present a dictionary learning method with an
effective control over the self-coherence of the trained dictionary,
enabling a trade-off between maximizing the sparsity of codings and
approximating an equiangular tight frame.\end{abstract}

\section{Introduction\label{sec:Introduction}}

\emph{Dictionary learning} adapts an initial dictionary to a particular
signal class with the help of training observations, such that further
observations from that class can be sparsely coded in the trained
dictionary with low approximation error. \emph{Over-complete} dictionaries,
consisting of more atoms than dimensions of the feature space, typically
support sparser codings by placing more atoms in densely populated
regions of the feature space. However, this redundancy increases the
\emph{self-coherence} of the dictionary, i.e.\ the pairwise similarity
of dictionary atoms, as measured by the cosine of the angle between
atom pairs. A lower self-coherence permits better \emph{support recovery}
\cite{donoho2006stability} and a more rapid decay of the residual
norm when increasing the coding cardinality \cite{Tropp2004}. Furthermore,
bounding the admissible self-coherence during training can increase
the generalization performance of the dictionary, by avoiding over-fitting
to the training data and by avoiding atom degeneracy, i.e. two atoms
collapsing onto the same vector.

We present a dictionary learning algorithm called IDL($\gamma$),
which enables an effective control over the self-coherence of \emph{trained}
dictionaries. Our method is able to span the full spectrum of optimization
objectives, from maximizing the sparsity of the resulting codings,
to approximating an \emph{equiangular tight frame} (ETF), which is
a dictionary achieving minimal self-coherence for a given number of
atoms. We demonstrate the benefits of limiting the self-coherence
of the dictionary in terms of better coding support recovery and improved
generalization performance (see Sec.~\ref{sec:Experiments}).

\subsection{From Bases to Over-Complete Dictionaries}

An orthonormal \emph{basis} $\mathbf{B}\in\mathbb{R}^{D\times D}$
contains $D$ mutually orthogonal unit $\ell_{2}$ norm atoms spanning
the feature space $\mathbb{R}^{D}$. The unique code $\mathbf{c}\in\mathbb{R}^{D}$
of an observation $\mathbf{x}\in\mathbb{R}^{D}$ is computed by $\mathbf{c}=\mathbf{B}^{\top}\mathbf{x}$
(\emph{signal analysis}), and the signal is recovered from the code
by $\mathbf{x}=\mathbf{B}\mathbf{c}$ (\emph{signal synthesis}). The
\emph{Gram} matrix $\mathbf{G}=\mathbf{B}^{\top}\mathbf{B}=\mathbf{I}$
of $\mathbf{B}$ is the identity matrix.

Although natural signals are approximately sparse in suitably chosen
bases, typically a sparser code can be achieved using an over-complete
dictionary $\mathbf{D}\in\mathbb{R}^{D\times L}$, with $L>D$ unit
$\ell_{2}$ norm atoms, by placing more atoms in densely populated
regions of the feature space. However, due to the redundant number
of atoms, coding $\mathbf{x}$ in $\mathbf{D}$ no longer has a unique
solution. Therefore, signal analysis in over-complete dictionaries
needs to be performed using a \emph{sparse coding algorithm}, such
as \emph{orthogonal matching pursuit} (OMP) \cite{mallat1993matching}.

The non-orthogonality of atoms is measured by the \emph{self-coherence}
of the dictionary, which can be defined as the maximum magnitude over
all off-diagonal elements of the Gram matrix $\mathbf{G}=\mathbf{D}^{\top}\mathbf{D}$,
\begin{equation}
\mu(\mathbf{D})=\max_{d,e}|(\mathbf{G}-\mathbf{I})_{(d,e)}|=\max_{d,e}|\mathbf{d}_{(:,d)}^{\top}\mathbf{d}_{(:,e)}|,\quad d\neq e.\label{eq:self_coherence}
\end{equation}
It therefore holds that $\mu(\mathbf{D})\in[0,1]$. Note that this
definition of the dictionary self-coherence can be misleading when
most inner products have small magnitudes \cite{Tropp2004}. Therefore,
the full inner product distribution is considered in Sec. \ref{sec:Experiments}.

$\mathbf{D}$ has minimum self-coherence\emph{ }for a given dimension
$D$ and dictionary size $L$\emph{ }if the magnitudes of all the
off-diagonal elements of $\mathbf{G}$ are equal (see Thm.~\ref{thm:-The-self-coherence}
below). In this case, the dictionary is called an \emph{equiangular
tight-frame} (ETF) \cite{Sustik2007}. Formally, $\mathbf{E}\in\mathbb{R}^{D\times L}$
 is an ETF if there exists an $\alpha$, $0<\alpha<\pi/2$, such that
\begin{equation}
|\mathbf{e}_{(:,d)}^{\top}\mathbf{e}_{(:,e)}|=\cos(\alpha),\quad d\neq e,
\end{equation}
and if
\begin{equation}
\mathbf{E}\mathbf{E}^{\top}=\frac{L}{D}\mathbf{I}.
\end{equation}

Therefore, $\mathbf{E}$ has $D$ non-zero singular values all equal
to $\sqrt{L/D}$. The following theorem establishes a lower bound
on the minimum of the self-coherence.
\begin{thm}
\label{thm:-The-self-coherence}\cite[Theorem 2.3]{Strohmer2003}
The self-coherence of a dictionary $\mathbf{D}\in\mathbb{R}^{D\times L}$
with unit $\ell_{2}$ norm atoms is bounded from below by
\begin{equation}
\mu(\mathbf{D})\geq\sqrt{\frac{L-D}{D(L-1)}}.
\end{equation}
Equality holds if and only if $\mathbf{D}$ is an ETF and $L\leq D(D+1)/2$.

\emph{The self-coherence of a dictionary influences the recovery of
the }sparse coding\emph{ }support\emph{ of a signal observation, i.e.
the set of atoms that are associated with the non-zero coding coefficients.
The }exact recovery condition \emph{(ERC) \cite{Donoho2001} states
that, assuming that the observation in fact has an exact sparse coding
$\tilde{\mathbf{c}}$ in $\mathbf{D}$, the support of $\tilde{\mathbf{c}}$
is recovered if} \textup{
\begin{equation}
\|\tilde{\mathbf{c}}\|_{0}<\frac{1}{2}\left(1+\frac{1}{\mu(\mathbf{D})}\right).\label{eq:ERC2}
\end{equation}
}\emph{Furthermore, $\mu(\mathbf{D})$ also upper bounds the residual
error norm decay curve in iterative sparse coding algorithms such
as OMP \cite{Tropp2004}.}
\end{thm}

\subsection{Related Work}

Yaghoobi et al.~proposed a design algorithm for parametric dictionaries
\cite{Yaghoobi2009}. A \emph{parametric dictionary} $\mathbf{D}_{\Gamma}$
consists of atoms which have a specific functional form controlled
by a small number of parameters. The proposed algorithm accepts a
given $\mathbf{D}_{\Gamma}$ as its input, and optimizes it such that
its Gram matrix approximates the optimal properties of an ETF. However,
this approach relies on expert knowledge for choosing the appropriate
parametric family for a given application, and provides no mechanism
to adapt $\mathbf{D}_{\Gamma}$ if the signal characteristics are
not known in advance. Therefore, an analytic dictionary design approach
is for instance not suited to source separation of partially coherent
sources \cite{Sigg2012}.

The K-SVD algorithm \cite{Aharon2006} adapts a non-parametric dictionary
to training data. In each iteration of the algorithm, those atoms
are replaced which have a too high coherence to another atom in the
dictionary. If the coherence to another atom lies above a threshold
$\mu_{t}$, the atom is replaced by a training observation which does
not have a sparse representation in the current dictionary. Therefore,
the likelihood that the replacement atom is less coherent to the dictionary
is high. However, if multiple atoms are replaced (which is almost
always the case in practice), this strategy does not guarantee that
the dictionary self-coherence falls below $\mu_{t}$. In our experiments,
an effective control over the dictionary self-coherence using the
proposed atom thresholding step was not possible (see Sec.~\ref{sec:Experiments}). 

Very recently, and independently from our own work, Mailh� et al.~\cite{Mailhe2012}
proposed a more sophisticated atom decorrelation step for the K-SVD
algorithm called INK-SVD, where pairs of atoms are decorrelated until
the dictionary satisfies the maximum inner product bound \eqref{eq:self_coherence}.
After the dictionary update step of the K-SVD algorithm is complete,
each pair of atoms which has a coherence above the threshold $\mu_{t}$
has its inner angle increased symmetrically until the threshold is
satisfied. Because this procedure can inadvertently increase the coherence
to other atoms, the pairwise decorrelation step has to be iterated
until the self-coherence threshold is satisfied for the complete dictionary.
Unfortunately, due to this fact the number of necessary decorrelation
steps can grow very large if a small $\mu_{t}$ is enforced (see Sec.
\ref{sec:Experiments}).

\subsection{Our Contribution}

We present a dictionary learning algorithm where a bound on the dictionary
self-coherence is enforced directly in the atom update step. Instead
of bounding the maximum inner product \eqref{eq:self_coherence} as
in the INK-SVD algorithm, our algorithm enforces an upper bound on
the sum of squared inner product values. By varying a Lagrange multiplier
$\gamma$, it is possible to realize any trade-off between maximizing
the sparsity of the code and minimizing the self-coherence of the
dictionary.

Since IDL($\gamma$) maximizes the coherence of a dictionary to a
particular signal class, prior expert knowledge to choose the right
parametric dictionary family and parameter discretization is not necessary.
Furthermore, the IDL($\gamma$) algorithm makes it possible to train
an incoherent dictionary even if the number of atoms is large compared
to the dimensionality of the signal space. And last but not least,
we empirically demonstrate for a speech coding task that training
an incoherent dictionary using IDL($\gamma$) improves the sparse
coding fidelity of the dictionary on unseen test data.

\section{Method\label{sec:Method}}

A dictionary learning algorithm approximately factorizes a data matrix
$\mathbf{X}\in\mathbb{R}^{D\times N}$ into a dictionary matrix $\mathbf{D}\in\mathbb{R}^{D\times L}$
and a coding matrix $\mathbf{C}\in\mathbb{R}^{L\times N}$. The algorithm
minimizes the approximation error 
\begin{equation}
\arg\min_{\mathbf{D},\mathbf{C}}\left\Vert \mathbf{X}-\mathbf{D}\cdot\mathbf{C}\right\Vert _{F}^{2},\label{eq:dl-equation}
\end{equation}
measured by the squared Frobenius norm, subject to a sparsity constraint
on \textbf{$\mathbf{C}$} and a unit $\ell_{2}$ norm constraint on
the atoms (columns) of $\mathbf{D}$. Since \eqref{eq:dl-equation}
is not jointly convex in $\mathbf{D}$ and $\mathbf{C}$, many proposed
algorithms employ alternating minimization w.r.t.\ $\mathbf{C}$
and $\mathbf{D}$ until convergence to a local optimum. In the following,
we focus our discussion on the dictionary update step.

The K-SVD algorithm minimizes \eqref{eq:dl-equation} for each atom
independently. Given the newly updated dictionary, if there exist
atoms $\mathbf{d}_{(:,d)}$ and $\mathbf{d}_{(:,e)}$, such that 
\begin{equation}
|\mathbf{d}_{(:,d)}^{\top}\mathbf{d}_{(:,e)}|>\mu_{t}\quad d\neq e
\end{equation}
$\mathbf{d}_{(:,e)}$ is replaced by $\mathbf{x}_{(:,n)}/\|\mathbf{x}_{(:,n)}\|_{2}$,
where $n$ is chosen such that $\|\mathbf{x}_{(:,n)}-\mathbf{D}\mathbf{c}_{(:,n)}\|_{2}$
is large. Since observations having a large approximation error are
likely incoherent to the current dictionary, the replacement atoms
likely have a coherence below $\mu_{t}$ to all atoms already in the
dictionary. However, if more than one atom is replaced, the coherence
between the replacement atoms can potentially be large. This approach
therefore does not guarantee that the self-coherence of the updated
dictionary falls below $\mu_{t}$.

Although updating atoms independently of each other is computationally
efficient, it is not well suited to enforcing a self-coherence constraint,
which introduces additional dependencies between all atoms. We propose
a dictionary update step where the atoms are jointly optimized, and
the dictionary self-coherence is minimized along with the approximation
error.

Thm. 1 motivates our choice to augment the minimization of the objective
\eqref{eq:dl-equation} w.r.t.\ $\mathbf{D}$ with a self-coherence
penalty,
\begin{equation}
\arg\min_{\mathbf{D}}\left\Vert \mathbf{X}-\mathbf{DC}\right\Vert _{F}^{2}+\gamma\|\mathbf{D}^{\top}\mathbf{D}-\mathbf{I}\|_{F}^{2}\label{eq:dl-self-coherence}
\end{equation}
where the Lagrange multiplier $\gamma$ controls the trade-off between
minimizing the approximation error and minimizing the self-coherence.
The second term in \eqref{eq:dl-self-coherence} penalizes both the
average coherence between atoms, as well as a divergence from the
unit $\ell_{2}$ norm of each atom. However, we still enforce the
strict unit $\ell_{2}$ norm constraint after the optimization by
rescaling each atom.

The gradient of \eqref{eq:dl-self-coherence} w.r.t.\ $\mathbf{D}$
is computed by a trace operator expansion, $\|\mathbf{A}\|_{F}^{2}=\mathrm{tr}\{\mathbf{A}^{\top}\mathbf{A}\}$,
of the approximation error term of \eqref{eq:dl-self-coherence},
\begin{equation}
\mbox{tr}\left\{ \mathbf{C}^{\top}\mathbf{D}^{\top}\mathbf{D}\mathbf{C}\right\} -2\mbox{tr}\left\{ \mathbf{X}^{\top}\mathbf{DC}\right\} +\mbox{tr}\left\{ \mathbf{X}^{\top}\mathbf{X}\right\} ,\label{eq:reconstruction-term}
\end{equation}
and the self-coherence penalty term of \eqref{eq:dl-self-coherence}
\begin{equation}
\mbox{tr}\left\{ \mathbf{D}^{\top}\mathbf{D}\mathbf{D}^{\top}\mathbf{D}\right\} -2\mbox{tr}\left\{ \mathbf{D}^{\top}\mathbf{D}\right\} +\mbox{tr}\left\{ \mathbf{I}\right\} .\label{eq:self-coherence-term}
\end{equation}
Taking the partial matrix derivatives of \eqref{eq:reconstruction-term}
and \eqref{eq:self-coherence-term} w.r.t.\textbf{\ $\mathbf{D}$
}results in the gradient
\begin{equation}
2\left(\mathbf{DCC}^{\top}-\mathbf{XC}^{\top}\right)+4\gamma\left(\mathbf{DD}^{\top}\mathbf{D}-\mathbf{D}\right),\label{eq:gradient}
\end{equation}
see e.g.~\cite{magnus1988matrix} how to take partial derivatives
of the trace operator.

It is not necessary to find the global minimizer of \eqref{eq:dl-self-coherence},
as long as the objective is sufficiently reduced in each iteration
of the dictionary learning algorithm. We therefore run only a few
iterations of the limited-memory BFGS algorithm \cite{Liu1989}, which
successively builds an approximation to the Hessian (i.e. the matrix
of second order partial derivatives) from evaluating the objective
\eqref{eq:dl-self-coherence} and the gradient \eqref{eq:gradient},
without directly computing the Hessian matrix (which is infeasible
for large dictionaries).

\section{Experiments\label{sec:Experiments}}

\begin{figure*}
\begin{centering}
\begin{tabular}{ccc}
\includegraphics[width=0.25\paperwidth]{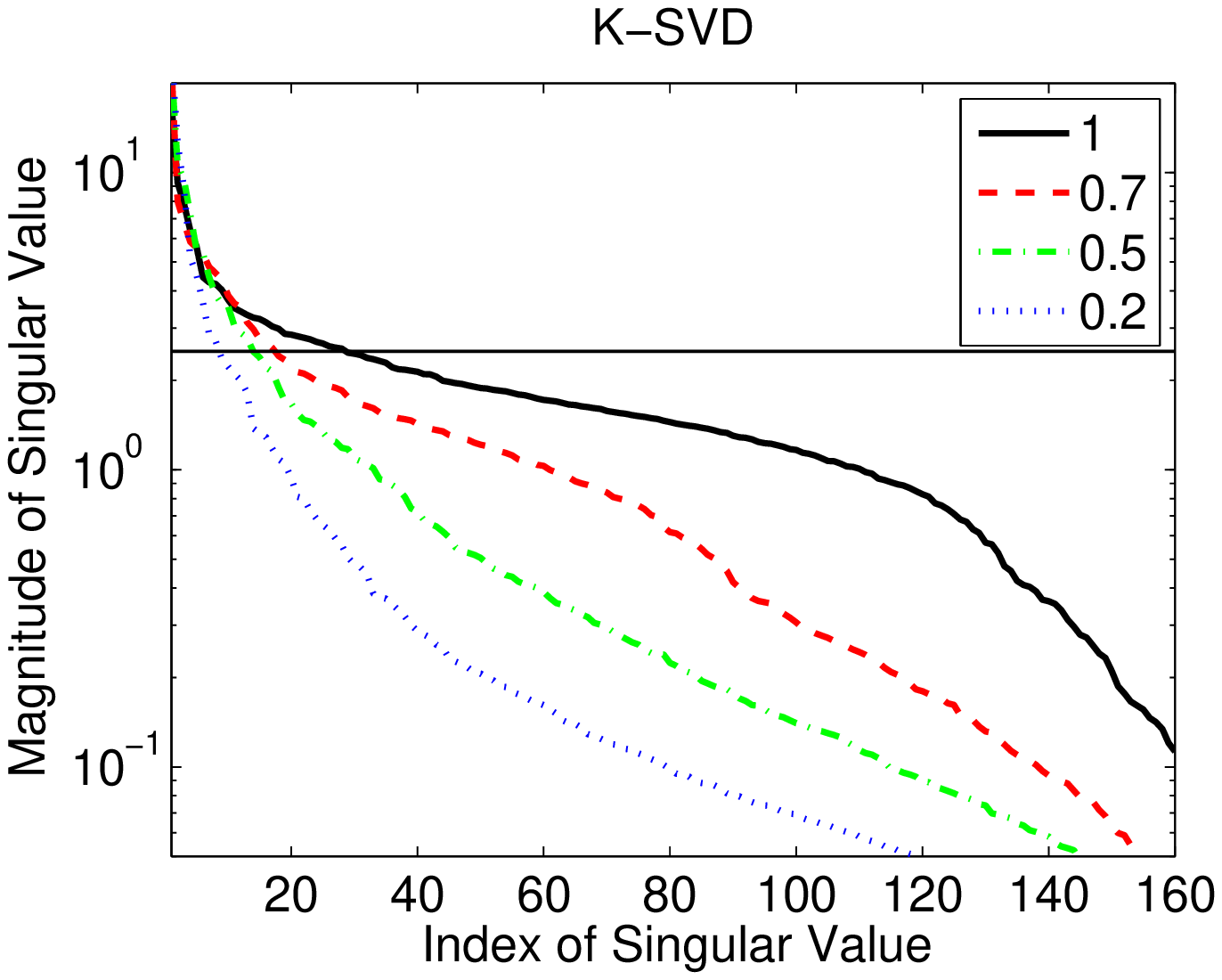} & \includegraphics[width=0.25\paperwidth]{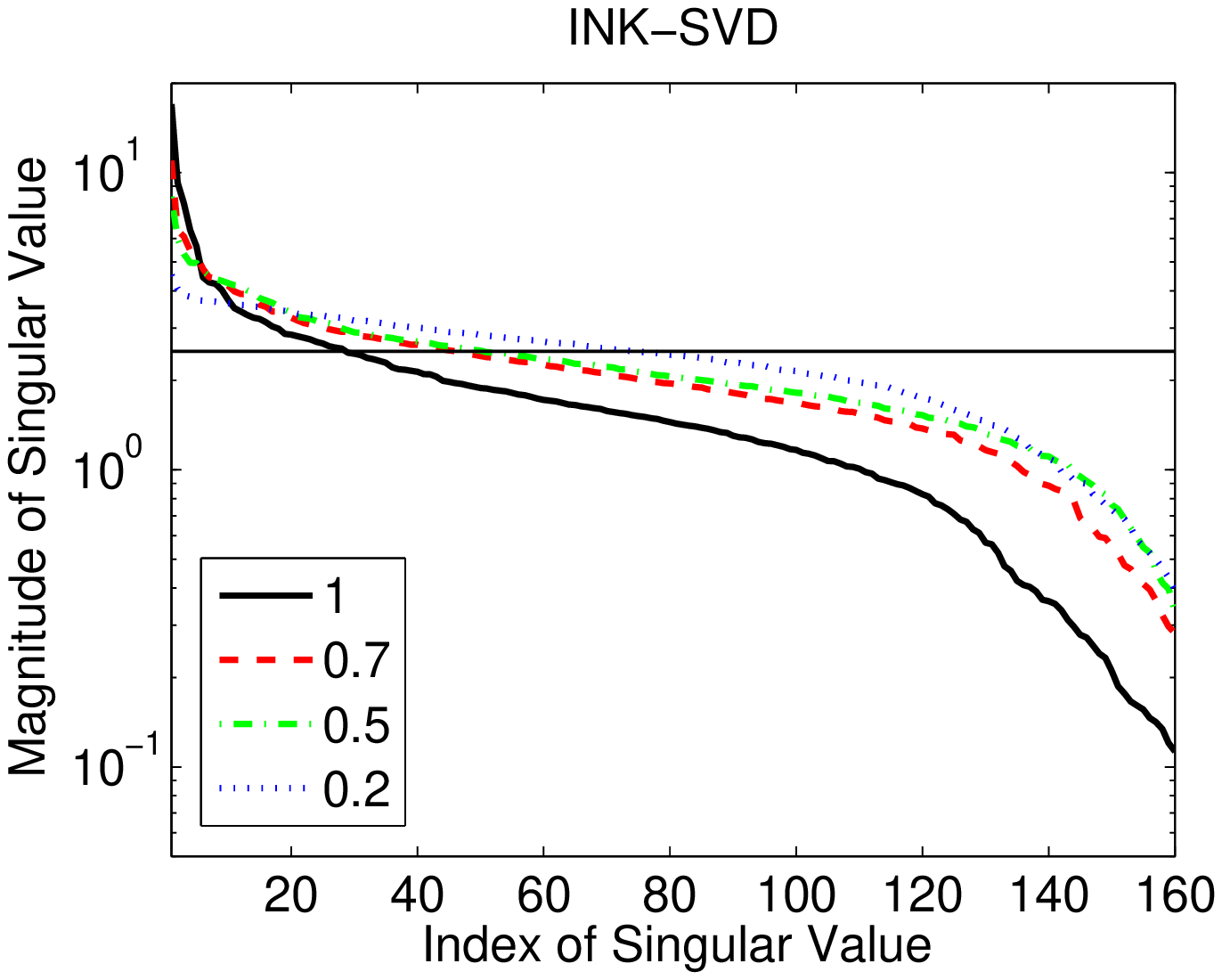} & \includegraphics[width=0.25\paperwidth]{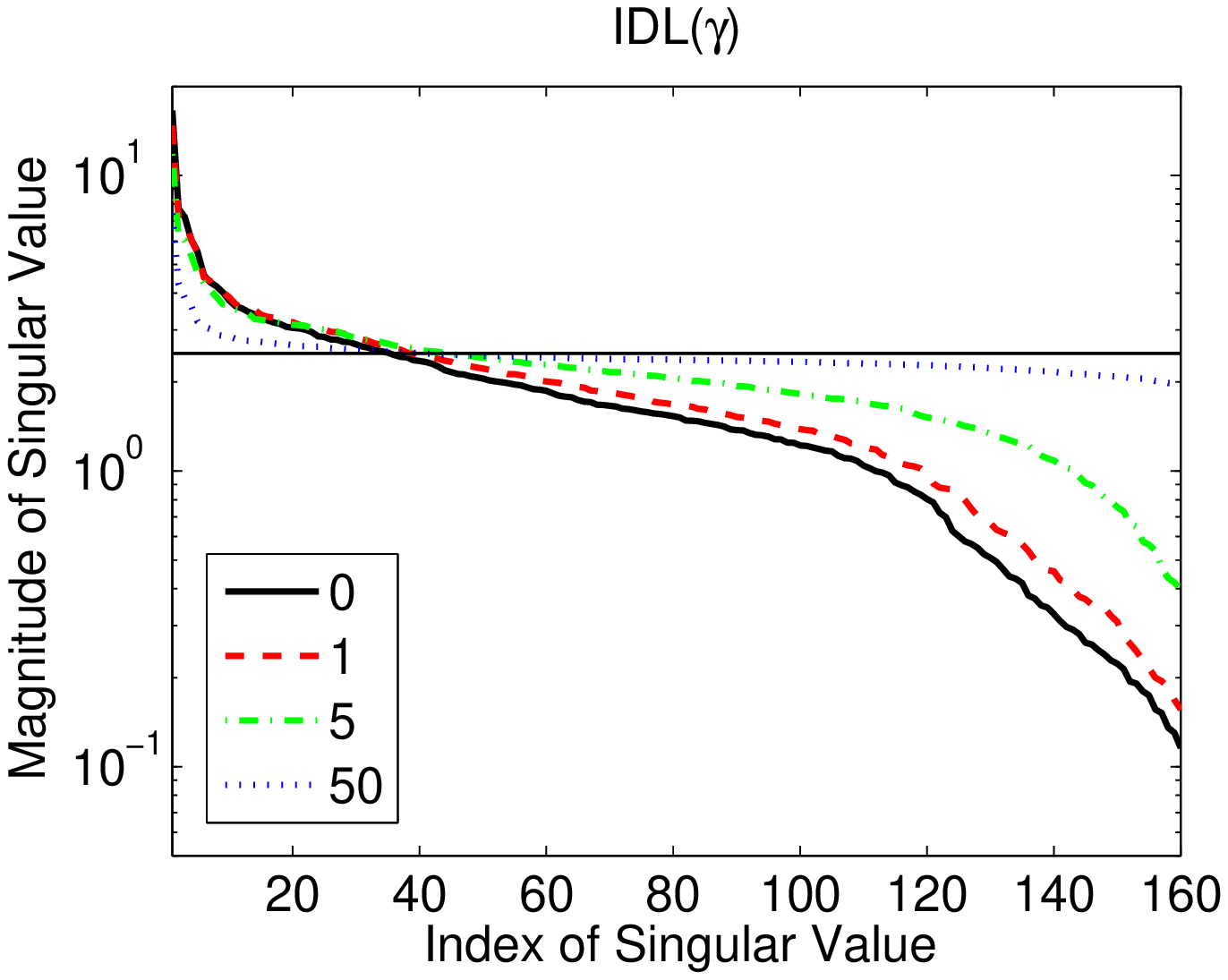}\tabularnewline
\end{tabular}
\par\end{centering}

\caption{\label{fig:dict-spec}Singular value spectra of the trained dictionaries,
as a function of the self-coherence constraint. A flatter spectrum
indicates a less coherent dictionary. As a reference, the constant
line indicates the flat spectrum of the corresponding ETF at $\sqrt{L/D}=2.5$.}
\end{figure*}
We compare the proposed dictionary learning algorithm, denoted IDL($\gamma$),
to the K-SVD algorithm with atom replacement and the INK-SVD algorithm.
The difference of our algorithm lies in the dictionary update: it
jointly minimizes both the data approximation error and the coherence
of all pairs of atoms. In contrast, the K-SVD and the INK-SVD algorithm
first perform a dictionary update step to minimize the data approximation
error, and then sequentially minimize the coherence of pairs of atoms.

The effectiveness of all algorithms to upper bound the dictionary
self-coherence was evaluated for a speech coding task, as follows.
The audio recordings of the first male speaker of the GRID%
\footnote{http://www.dcs.shef.ac.uk/spandh/gridcorpus/%
} corpus were randomly sub-sampled to obtain $N=30000$ training signals,
each $D=160$ samples long. A dictionary with $L=1000$ atoms was
initialized using random sampling of training observations. The LARC
algorithm \cite{Sigg2012} (an extension of the LARS algorithm \cite{Efron2004})
was used for the sparse coding step of all dictionary learning algorithms,
with the LARC residual coherence threshold set to $\mu_{\mathrm{dl}}=0.2$
(not to be confused with the self coherence threshold $\mu_{t}$).
The number of dictionary learning iterations was set to 25, which
resulted in approximate convergence to a local optimum in all experiments. 

Figure \ref{fig:dict-spec} plots the singular value spectra of the
trained dictionaries. As a reference, the constant line at $\sqrt{L/D}=2.5$
indicates the flat spectrum of a corresponding ETF. For the K-SVD
algorithm (left figure), setting $\mu_{t}=1$ implies that the upper
bound on the self-coherence is inactive. Note that decreasing $\mu_{t}$
below unity proved to be counterproductive, i.e. the singular value
spectrum decreases even more rapidly. As desired, lowering $\mu_{t}$
for the INK-SVD algorithm resulted in a flatter spectrum (middle figure),
but the computational cost is increasingly dominated by the growing
number of decorrelation steps. Thus we were unable to train a dictionary
with $\mu_{t}=0.1$ (or smaller) in the available time frame (24 hours
on an Intel Core 2 Duo CPU). The results for IDL($\gamma$) (right
figure) show that by increasing the influence of the self-coherence
penalty in \eqref{eq:dl-self-coherence}, it is possible to approximate
the flat spectrum of an ETF. Setting $\gamma>50$ resulted in even
flatter spectra (not shown). Atom coherence histograms and atom recovery
percentages are available from the paper companion webpage%
\footnote{http://sigg-iten.ch/research/spl2012/%
}.

\begin{figure*}
\begin{centering}
\begin{tabular}{ccc}
\includegraphics[width=0.25\paperwidth]{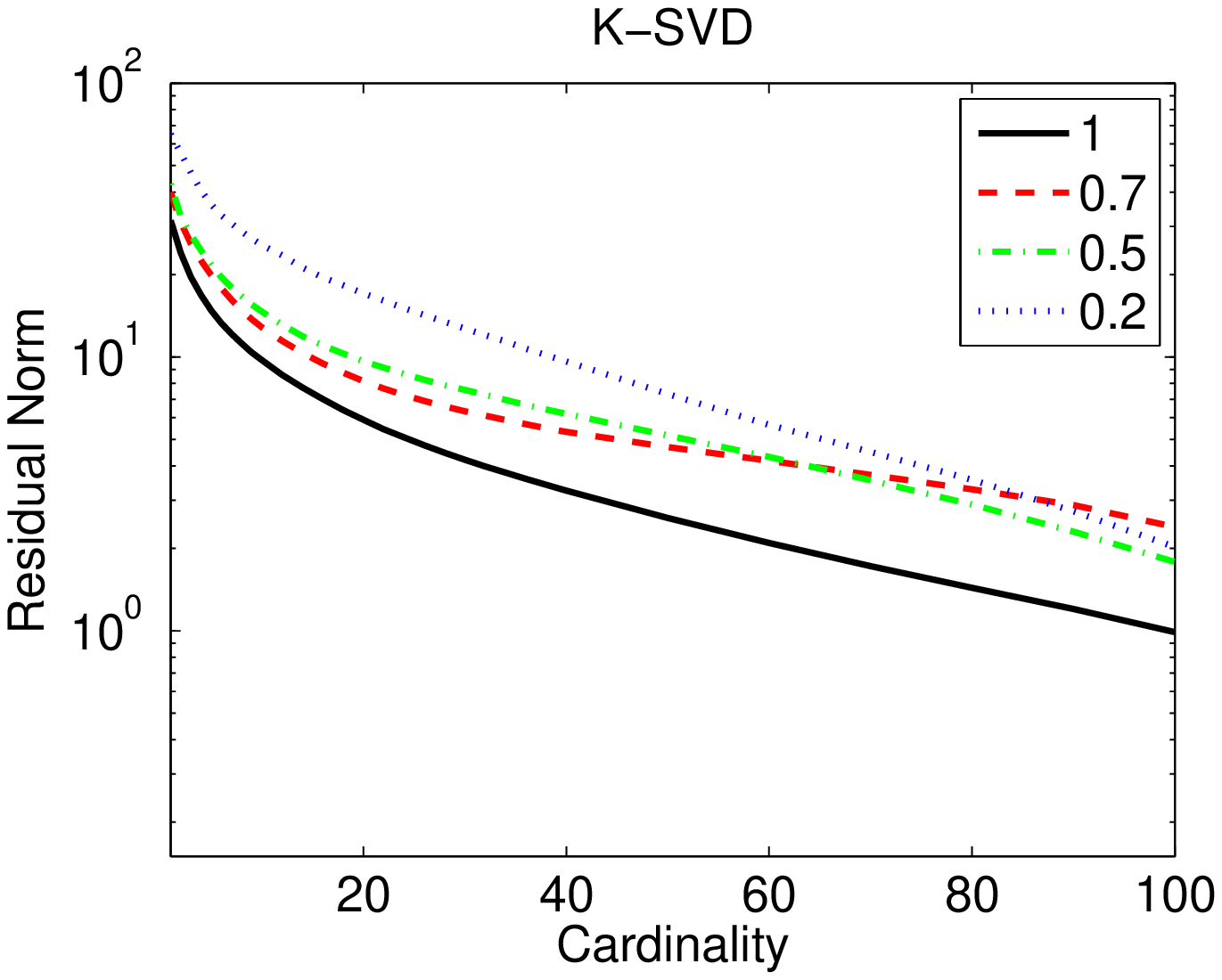} & \includegraphics[width=0.25\paperwidth]{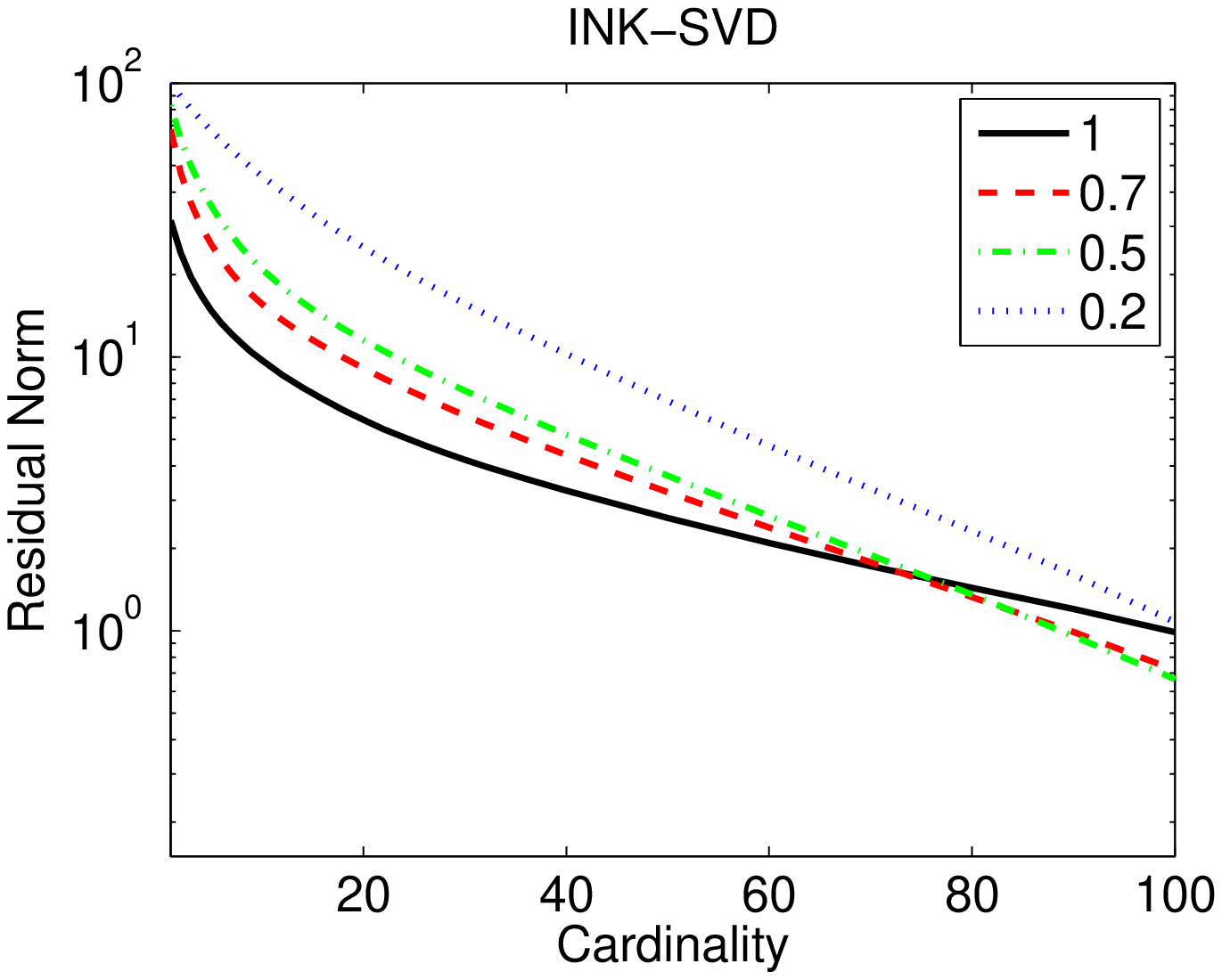} & \includegraphics[width=0.25\paperwidth]{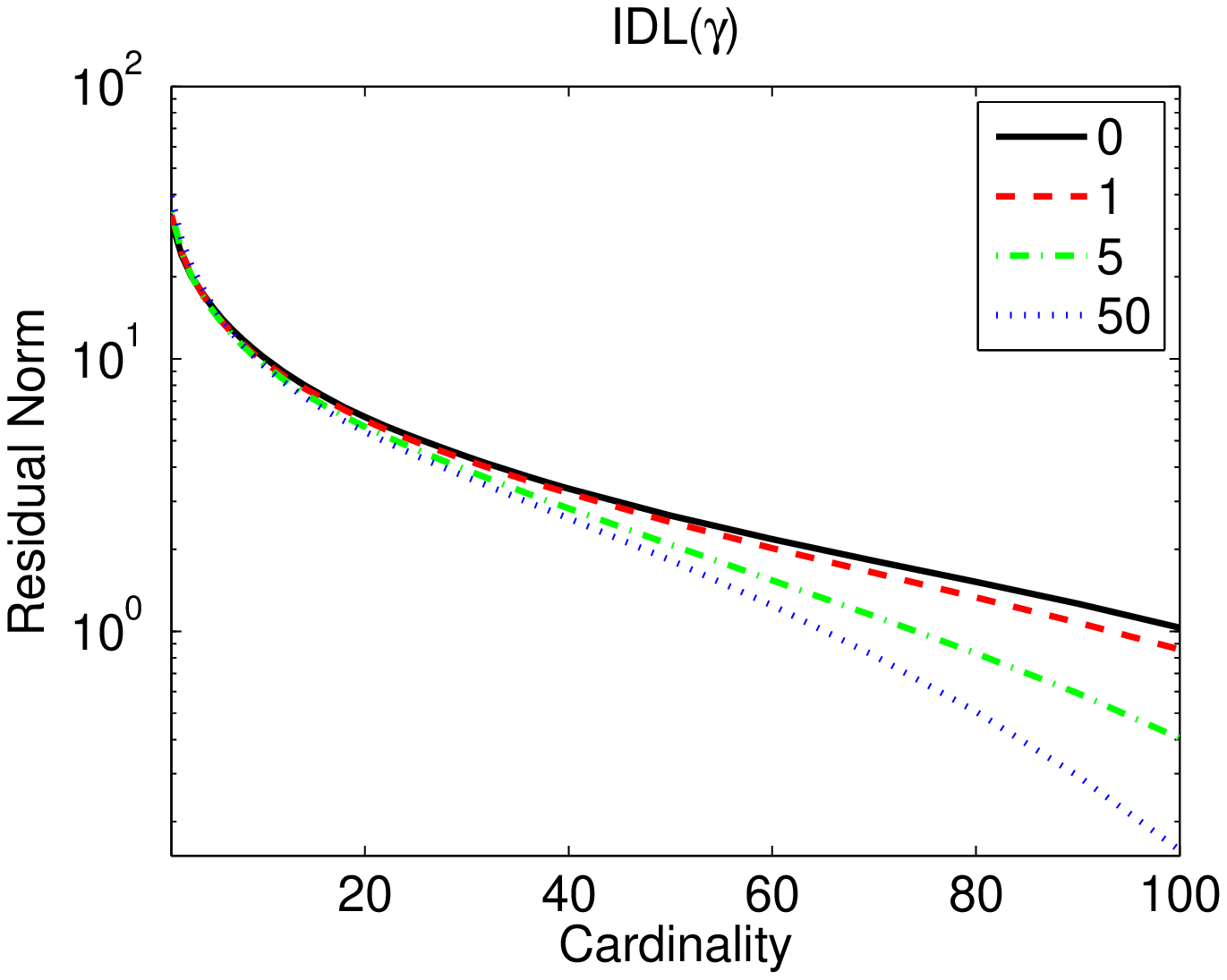}\tabularnewline
\end{tabular}
\par\end{centering}

\caption{\label{fig: dict-generalize}Generalization performance of the trained
dictionaries, as a function of the self-coherence constraint. Smaller
values indicate a better trade-off between the residual norm and the
coding cardinality on test data not seen during training.}
\end{figure*}

Figure \ref{fig: dict-generalize} plots the generalization performance
of the trained dictionaries, in terms of the trade-off between the
residual norm and the cardinality of the coding. Twenty test utterances
were coded using OMP with a cardinality stopping criterion, and the
median residual norm is reported. For the K-SVD algorithm, decreasing
$\mu_{t}<1$ resulted in a deteriorating generalization performance.
For the INK-SVD algorithm, decreasing the residual norm is possible
for $0.7>\mu_{t}>0.2$ at cardinalities beyond 80, but only at the
cost of increasing the residual norm at smaller cardinalities. While
the curves are nearly identical for all algorithms if no coherence
bound is enforced, the generalization performance improves consistently
only in the case of IDL($\gamma$). We conjecture that the difference
is due to joint minimization of the residual norm and the dictionary
self-coherence in IDL($\gamma$), whereas the atom decorrelation of
K-SVD and INK-SVD is independent of the dictionary update.

\section{Conclusions and Discussion}

We present a dictionary learning algorithm which enables an effective
control over the self-coherence of the trained dictionary, enabling
a trade-off between maximizing the sparsity of the code and approximating
an equiangular tight frame. Neither a simple replacement of too similar
atoms or a pairwise decorrelation of atoms can both effectively and
efficiently control the dictionary self-coherence. We propose a joint
atom update step instead, simultaneously minimizing the approximation
error and the coherence of all pairs of atoms.

We show for a speech coding task that our method is able to achieve
the full range of optimization objectives, from maximizing the coding
sparsity to approximating the properties of an ETF. Furthermore, we
demonstrate the benefits of bounding the dictionary self-coherence
on the generalization performance of the dictionary.

\section*{Acknowledgments}

We thank Dirk-Jan Kroon for publishing an excellent Matlab implementation
of L-BFGS%
\footnote{http://www.sas.el.utwente.nl/open/people/Dirk-Jan\%20Kroon%
}. We also thank the reviewers for their valuable feedback and for
drawing our attention to \cite{Mailhe2012}.

\bibliography{bibliography}

\end{document}